\PassOptionsToPackage{bookmarks={false}}{hyperref}
\documentclass[10pt,twocolumn,letterpaper]{article}

\usepackage{wacv}
\usepackage{times}
\usepackage{epsfig}
\usepackage{graphicx}
\usepackage{amsmath}
\usepackage{amssymb}
\usepackage{gensymb}
\usepackage{multirow}
\usepackage{rotating}
\usepackage{fancyhdr}
\usepackage{latexsym}
\usepackage{url}
\usepackage{bm}
\usepackage{stmaryrd}
\usepackage{dsfont}
\usepackage{epstopdf}
\usepackage{booktabs}
\usepackage{mathalfa}
\usepackage{tabularx}
\usepackage{color}
\usepackage{caption}
\usepackage{enumitem}
\setlist{nosep}
\usepackage[ruled, lined, longend, linesnumbered]{algorithm2e}

\AppendGraphicsExtensions{.tiff}

\usepackage{caption}
\usepackage{subcaption}

\usepackage{xspace}
\makeatletter
\DeclareRobustCommand\onedot{\futurelet\@let@token\@onedot}
\def\@onedot{\ifx\@let@token.\else.\null\fi\xspace}
\def\eg{\emph{e.g}\onedot} 
\def\ie{\emph{i.e}\onedot}

\def\etal{\emph{et~al}\onedot}
\makeatother

\wacvfinalcopy 

\def\wacvPaperID{348} 

\ifwacvfinal\fi
\begin{document}

\addtolength{\baselineskip}{-0.3pt}

\title{Semantic Instance Meets Salient Object: \\Study on Video Semantic Salient Instance Segmentation\thanks{This work was in part supported by CART of the Ministry of Land, Infrastructure, Transport and Tourism of Japan,
JST CREST (Grant No. JPMJCR14D1), and Grant-in-Aid for Scientific Research (Grant No. 16H02851) of the Ministry of Education, Culture, Sports, Science, and Technology of Japan.
}}


\author{
Trung-Nghia Le\thanks{Trung-Nghia Le is now with Kyushu University, Japan.} 
\\{\tt\small le.trung.nghia.437@m.kyushu-u.ac.jp}
\\{\small Graduate University for Advanced Studies (SOKENDAI), Japan}
\and
Akihiro Sugimoto
\\{\tt\small sugimoto@nii.ac.jp}
\\{\small National Institute of Informatics, Japan}
}

\maketitle
\ifwacvfinal\thispagestyle{empty}\fi

\begin{abstract}

Focusing on only semantic instances that only salient in a scene gains more benefits for robot navigation and self-driving cars than looking at all objects in the whole scene. This paper pushes the envelope on salient regions in a video to decompose them into semantically meaningful components, namely, semantic salient instances. We provide the baseline for the new task of video semantic salient instance segmentation (VSSIS), that is, Semantic Instance - Salient Object (SISO) framework. The SISO framework is simple yet efficient, leveraging advantages of two different segmentation tasks, \ie~ semantic instance segmentation and salient object segmentation to eventually fuse them for the final result. In SISO, we introduce a sequential fusion by looking at overlapping pixels between semantic instances and salient regions to have non-overlapping instances one by one. We also introduce a recurrent instance propagation to refine the shapes and semantic meanings of instances, and an identity tracking to maintain both the identity and the semantic meaning of instances over the entire video. Experimental results demonstrated the effectiveness of our SISO baseline, which can handle occlusions in videos. In addition, to tackle the task of VSSIS, we augment the DAVIS-2017 benchmark dataset by assigning semantic ground-truth for salient instance labels, obtaining SEmantic Salient Instance Video (SESIV) dataset. Our SESIV dataset consists of 84 high-quality video sequences with pixel-wisely per-frame ground-truth labels. 

\end{abstract}



\section{Introduction}
\label{c5_section:introduction}

\begin{figure}[t]
    \centering
        \includegraphics[width=1\linewidth]{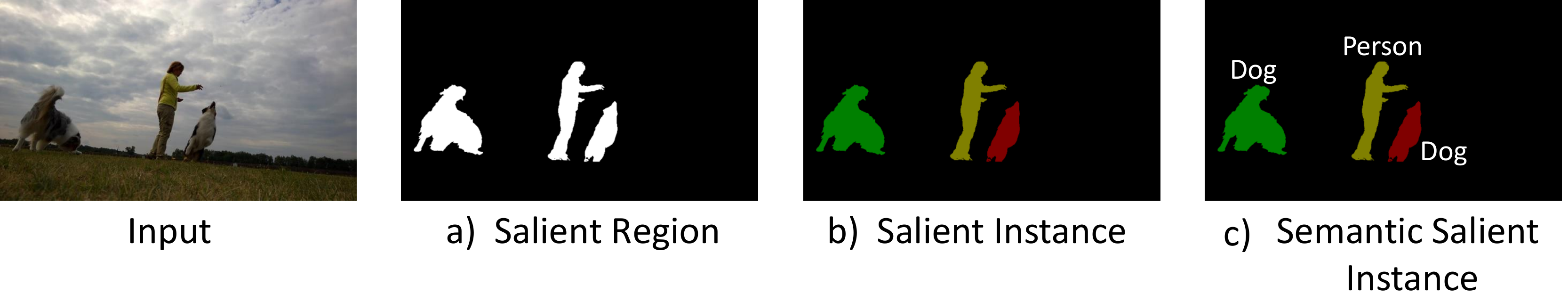}
       \vspace*{-1.5\baselineskip}
    \caption{Segmentation levels of salient objects. The input video frame is followed by different levels of label annotation. Our work focuses on segmenting semantic salient instances (most right).}
    \label{c5_fig:saliency_tasks}
\end{figure}

Recent advances in salient object segmentation (SOS) in videos using CNN~\cite{ltnghia-BMVC2017, Li-CVPR2018, Li-TIP2018, Wang-TIP2018} have demonstrated impressive performance in accuracy. 
Such SOS methods~\cite{ltnghia-BMVC2017, Li-CVPR2018, Li-TIP2018, Wang-TIP2018} focus on only localizing the region of interest by labeling ``salient" or ``non-salient" to each pixel in the video frame. The localized salient region, however, may involve multiple (interacting) objects (Fig.~\ref{c5_fig:saliency_tasks} a), which is a more reasonable scenario in the real-world scenes.
Therefore, localized salient regions should be decomposed into conceptually meaningful components (Fig.~\ref{c5_fig:saliency_tasks} b), called salient instances~\cite{Li-CVPR2017}, for better understanding of videos.
Furthermore, attaching a semantic label to each salient instance (Fig.~\ref{c5_fig:saliency_tasks} c)
will widen the range of applications of SOS even to 
autonomous driving~\cite{Zhang-CVPR2016} and robotic interaction~\cite{Xu-CVPR2016}.
Nevertheless, segmenting semantic salient instances is not yet addressed in the literature. 




\begin{figure}[t]
    \includegraphics[width=1\linewidth]{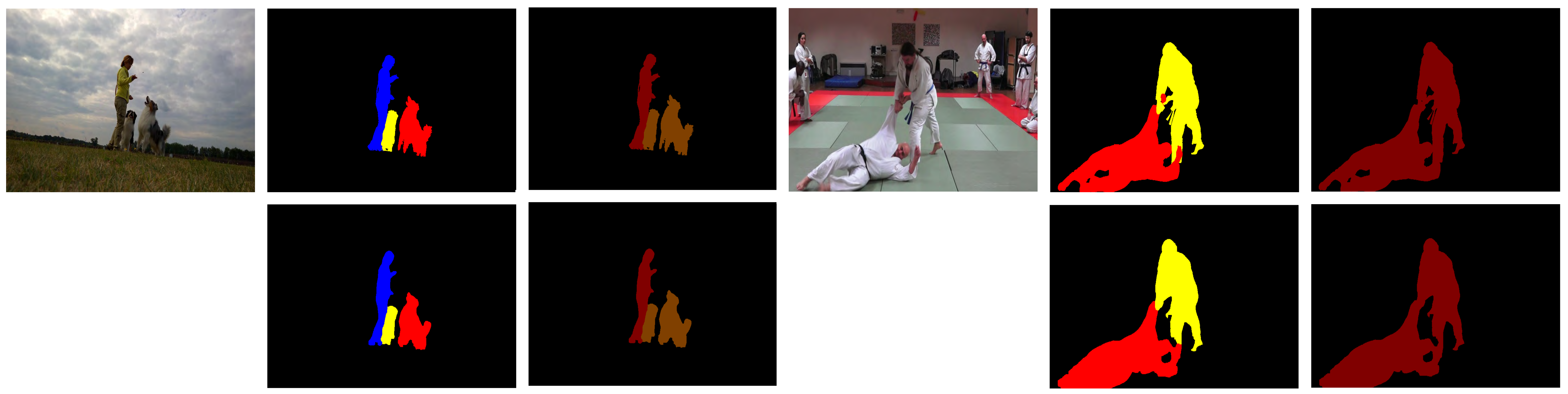}
    \vspace*{-\baselineskip}
    \caption{Examples obtained by our method on the SESIV dataset. From left to right, the original video frame is followed by instance label and semantic label. The first and second rows show ground-truth labels, and segmented results, respectively.}
    \label{c5_fig:examples}
\end{figure}

To achieve this semantic-instance level segmentation of salient regions, we aim to jointly identify individual instances in the segmented salient regions and categorize these salient instances (Fig.~\ref{c5_fig:saliency_tasks} c). We refer this problem to \textit{semantic salient instance segmentation}, which aims to identify \textit{only individual prominent foreground object} classes. 
The problem is even more challenging on the videos because instances need to be tracked over the entire video to maintain their identifications even if they are occluded at some frames. We remark that in this paper, an instance in a video is defined to be salient if it appears in the first video frame and stands out for more than 50\% duration of the video in total.




Many computer vision tasks such as action localization~\cite{He-WACV2018}, action recognition~\cite{Choutas-CVPR2018}, object relation detection~\cite{Hu-CVPR2018}, or video captioning~\cite{Wang-CVPR2018}, focus on dominant objects in the scene to avoid the expensive computational cost. 
Narrowing down dominant objects further using semantic salient instances is more appropriate in real application scenarios.  
Indeed, for autonomous robots or self-driving car, it is sufficient to focus on only a few useful semantic instances on the street view such as pedestrians or cars with high performance in accuracy and processing time instead of looking at all semantic object classes in the whole scene.

There are many methods proposed for each task of semantic instance segmentation (SIS)~\cite{Dai-CVPR2016, Kaiming-ICCV2017, Liu-CVPR2018} and SOS~\cite{Li-TIP2018, Liu-CVPR2016, Wang-TIP2018}. However, to the best of our knowledge, no work exists on semantic salient instance segmentation in images or videos to the date. Li \etal~\cite{Li-CVPR2017} very recently proposed a method for salient instance segmentation for images but do not deal with semantic level of segmentation.

On the other hand, the CNN-based approach to SOS requires a large number of training samples.  As illustrated in 
Table \ref{c5_tab:SOS_dataset}, several benchmark datasets for various tasks of SOS have been provided~\cite{Borji-TIP2015, Brox-ECCV2010, Cheng-TCV2014, Cheng-CVPR2011, Fukuchi-ICME2009, Li-CVPR2015,  Li-TIP2018, Perazzi-CVPR2016, Tsai-IJCV2012, Wang-CVPR2017, Wang-TIP2015, Xia-CVPR2017, Yan-CVPR2013}.
The dataset quality is improved over the time in terms of the number of samples and the detailed annotation. 
Though some datasets for salient instance segmentation are recently available (\eg~ SOI dataset~\cite{Li-CVPR2017} for images and SegTrack2 dataset~\cite{Li-ICCV2013} for videos),
they do not have sufficient numbers of samples to train deep networks. 
For semantic salient instance segmentation, to the best of our knowledge, no dataset having a sufficient number of samples for training is available to the date.

The overall contribution of this paper is three-fold:

First, we address the new task of \textbf{video semantic salient instance segmentation (VSSIS)} and analyze in-depth challenges of the problem. Finding semantic salient instances in videos is a useful task and it can be an interesting problem for the community. Existing work individually performs SIS or SOS, but no work can jointly perform these two tasks, which is considered as the new task of VSSIS.

Second, we introduce the baseline for VSSIS, called \textbf{S}emantic \textbf{I}nstance - \textbf{S}alient \textbf{O}bject (\textbf{SISO}). SISO is a simple yet efficient two-stream framework leveraging advantages of two different segmentation tasks, \ie~ SIS and SOS, through combining outputs of two streams. 
SISO possesses three key features: sequential fusion, recurrent instance propagation, and identity tracking. 
The sequential fusion frame-wisely fuses the outputs of the two streams. Using our introduced instance merging order and frame-confidence, the salient region obtained from the SOS stream is decomposed into non-overlapping salient instances one by one. 
The recurrent instance propagation recovers unsegmented semantic salient instances by recurrently propagating instances in frames with high frame-confidence to ones in frames with low frame-confidence. 
Identity tracking, on the other hand, maintains the consistency of instance identities and semantic labels over the entire video where identity propagation is for short-term consistency and re-identification is for long-term consistency. We also comprehensively evaluate the performance of the proposed baseline and deeply analyze results to show promising avenues for future research. 


Third, we provide a dataset, \textbf{SE}mantic \textbf{S}alient \textbf{I}nstance \textbf{V}ideo (\textbf{SESIV}) dataset \footnote{The SESIV annotations and evaluation scripts are publicly available at \textcolor{red}{https://sites.google.com/view/ltnghia/research/sesiv}}
accompanied with complementary metrics specifically designed for the task of VSSIS. The SESIV dataset consists of 84 high-quality video sequences with various densely annotated, pixel-accurate and per-frame ground-truth labels for different segmentation tasks. Our SESIV annotations are built on top of existing DAVIS-2017 annotations~\cite{Jordi-2017}. From pixel-wise instance-level labels of the DAVIS-2017 dataset, we identify salient instances and assign a semantic label to each instance. We emphasize that this is the very first dataset for VSSIS. Figure \ref{c5_fig:examples} shows some example results obtained by the SISO baseline in the SESIV dataset. We believe that our introduced SESIV dataset and metrics raise interest to the community and promote further research on VSSIS. 

\begin{table}[t]
\centering
\caption{Datasets for salient object segmentation tasks.}
\vspace*{-0.5\baselineskip}
\label{c5_tab:SOS_dataset}
\resizebox{\linewidth}{!}{%
\begin{tabular}{l|ll}
\toprule
\textbf{Task}    & \centering \textbf{Image}     & \textbf{Video}      \\ \midrule


\textbf{\begin{tabular}[c]{@{}l@{}}Salient Object\\ Segmentation\end{tabular}}            & \begin{tabular}[c]{@{}l@{}}MSRA~\cite{Cheng-CVPR2011}, CSSD~\cite{Yan-CVPR2013},\\ Judd-A~\cite{Borji-TIP2015}, THUR~\cite{Cheng-TCV2014},\\ HKU-IS~\cite{Li-CVPR2015}, XPIE~\cite{Xia-CVPR2017},\\ DUTS~\cite{Wang-CVPR2017}\end{tabular}  & \begin{tabular}[c]{@{}l@{}} SegTrack~\cite{Tsai-IJCV2012},\\ DAVIS-2016~\cite{Perazzi-CVPR2016},\\ 10-Clips~\cite{Fukuchi-ICME2009}, FBMS~\cite{Brox-ECCV2010},\\ ViSal~\cite{Wang-TIP2015}, VOS~\cite{Li-TIP2018}\end{tabular} \\ \midrule

\textbf{\begin{tabular}[c]{@{}l@{}}Salient Instance\\ Segmentation\end{tabular}}          & \begin{tabular}[c]{@{}l@{}}
SOI~\cite{Li-CVPR2017}\end{tabular}   & SegTrack2~\cite{Li-ICCV2013}   \\ \midrule

\begin{tabular}[c]{@{}l@{}}\textbf{Semantic Salient}\\ 
\textbf{Instance Segmentation}\end{tabular} & None & \textbf{Our proposed SESIV}    \\             \bottomrule                                                          
\end{tabular}
}
\end{table}

\section{Related Work}
\label{c5_section:related_work}


Semantic instance segmentation (SIS) is the task of unifying object detection and semantic segmentation. It has been intensively studied in recent years where the segmentation based approach or the proposal based approach is employed. The segmentation based approach~\cite{Kirillov-CVPR2017, Levinkov-CVPR2017, Liang-TPAMI2017, Zhang-CVPR2016} generally adopts two-stage processing: segmentation first and then instance clustering. The proposal based approach~\cite{Chen-CVPR2018, Dai-CVPR2016, Kaiming-ICCV2017, Yi-CVPR2017, Liu-CVPR2018}, on the other hand, predicts bounding-boxes first and then parses the bounding-boxes to obtain mask regions~\cite{Dai-CVPR2016} or exploits object detection models (\eg, Faster R-CNN~\cite{Ren-NIPS2015} or R-FCN~\cite{Dai-NIPS2016}) to classify mask regions~\cite{Chen-CVPR2018, Kaiming-ICCV2017, Yi-CVPR2017, Liu-CVPR2018}. Among these methods, Mask R-CNN~\cite{Kaiming-ICCV2017}
achieves the state-of-the-art performance, and recent work~\cite{Girdhar-CVPR2018, Liu-CVPR2018} is based on Mask R-CNN's architecture. To the best of our knowledge, no work exists that deals with video semantic instance segmentation. We thus use the strategy of frame-by-frame segmentation followed by instance linkage over the entire video.



\begin{figure*}[t]
    \centering
        \includegraphics[width=1\textwidth]{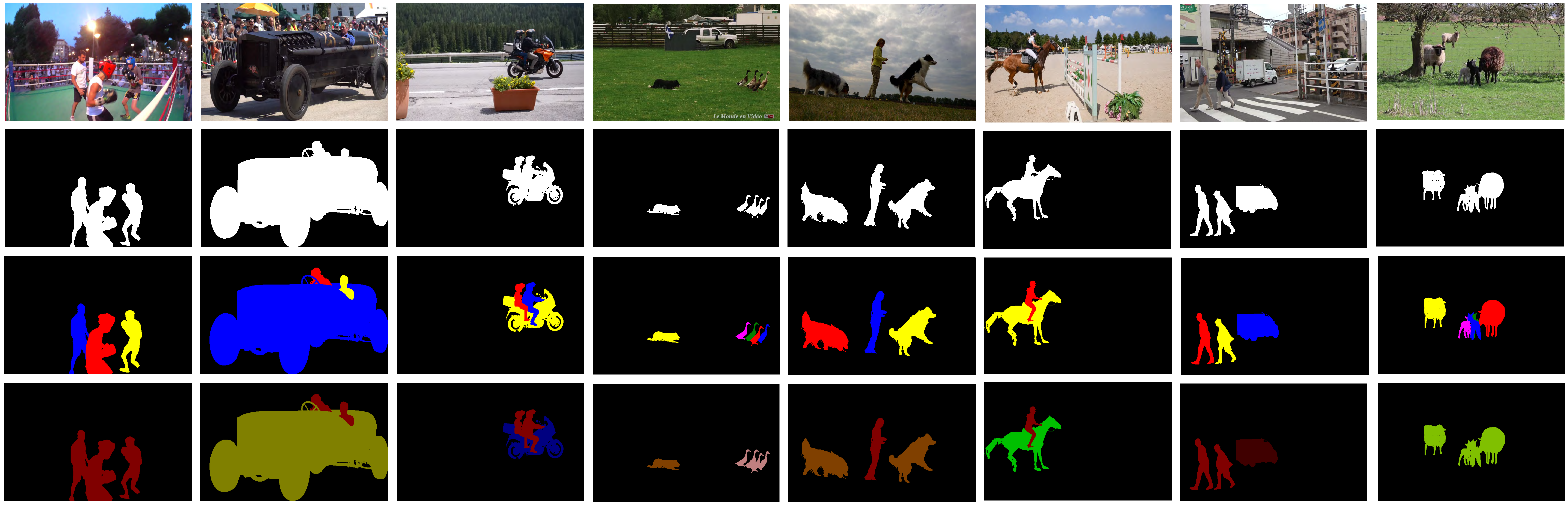}
    \caption{Samples from the SESIV dataset. From top to bottom, video frame is followed by saliency ground-truth, instance ground-truth, and semantic label ground-truth.}
    \label{c5_fig:SESIV_dataset2}
\end{figure*}

\begin{figure*}[t]
    \centering
        \includegraphics[width=1\textwidth]{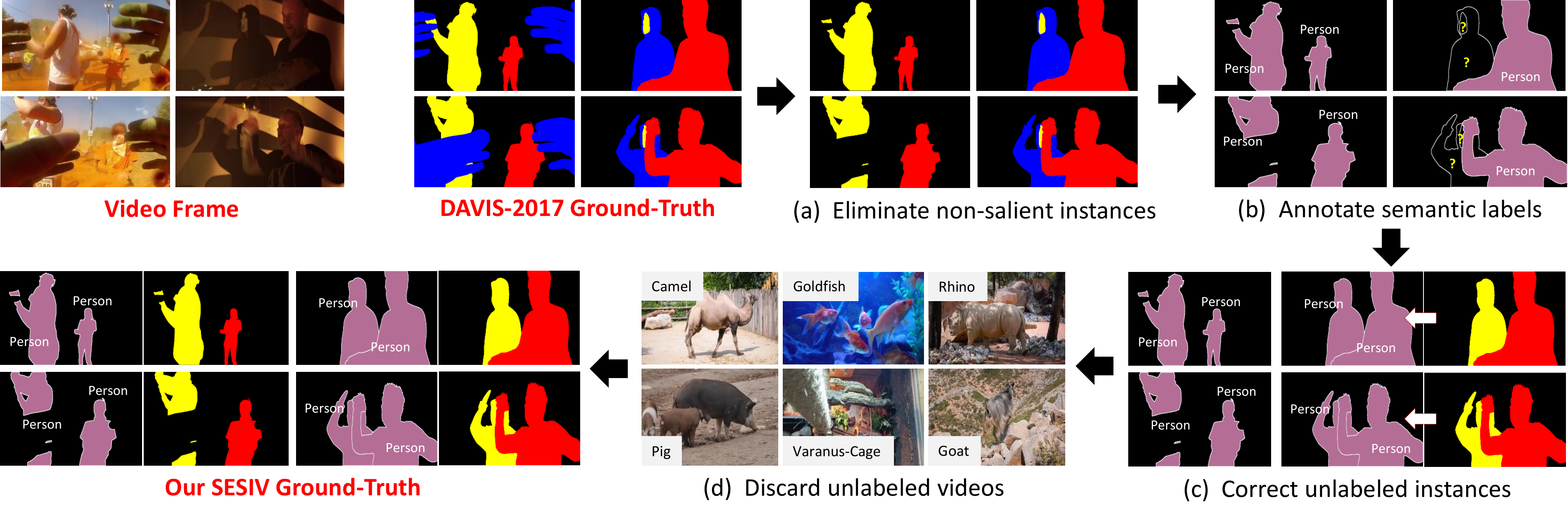}
    \vspace*{-1.5\baselineskip} 
    \caption{SESIV dataset construction.}
    \vspace*{-0.5\baselineskip}
    \label{c5_fig:dataset_construction}
\end{figure*}

Recent video salient object segmentation (VSOS) methods are based on the convolutional neural network (CNN)~\cite{ltnghia-BMVC2017, ltnghia-DeLIMMA2017, Li-CVPR2018, Li-TIP2018, Wang-TIP2018} and have demonstrated superior results over early work utilizing only hand-crafted features~\cite{ltnghia-PSIVT2015, Liu-PAMI2011, Rahtu-ECCV2010, Wang-CVPR2015, Wang-TIP2015, Zhou-CVPR2014}. 
These CNN based methods are classified into two approaches: segmentation based approach and end-to-end saliency inference approach. The segmentation based approach first segments each frame of a video into regions and uses deep features extracted from each region for saliency inference~\cite{ltnghia-DeLIMMA2017}. The end-to-end saliency inference approach, on the other hand, uses fully convolutional networks (FCNs)~\cite{ltnghia-BMVC2017, Li-CVPR2018, Li-TIP2018, Wang-TIP2018} to utilize optical flow~\cite{Li-CVPR2018, Li-TIP2018, Wang-TIP2018} or 3D kernels~\cite{ltnghia-BMVC2017}. The end-to-end saliency inference approach achieves better performance than the segmentation based one, and using 3D kernels can deal with more frames than optical flow to incorporate temporal information. We thus employ \cite{ltnghia-BMVC2017} as the SOS stream in SISO.



\section{Semantic Salient Instance Video Dataset}


\subsection{Overview}


To promote VSSIS, a publicly available dataset with pixel-wise ground-truth annotation is mandatory. 
We thus construct the \textbf{SE}mantic \textbf{S}alient \textbf{I}nstance \textbf{V}ideo (SESIV) dataset. We emphasize that no other dataset is publicly available for VSSIS. Figure \ref{c5_fig:SESIV_dataset2} illustrates examples from our SESIV dataset with their corresponding ground-truth labels.

The proposed SESIV dataset consists of 84 videos with 185 semantic salient instances categorized into 29 classes. The training set consists of 58 videos (with 136 instances and 27 categories), and the testing set consists of 26 videos (with 49 instances and 14 categories). For each video frame, we provide various ground-truth labels (\ie, saliency label, instance label, and semantic label, as exampled in Fig.~\ref{c5_fig:SESIV_dataset2}). We remark that SESIV annotations are built on top instance-level ground-truth labels of the DAVIS-2017 dataset~\cite{Jordi-2017}.

\subsection{Dataset Construction}

\begin{figure*}[t]
    \centering
    \begin{tabularx}{\textwidth}{*{3}{X}}
        \centering \includegraphics[width=1\linewidth]{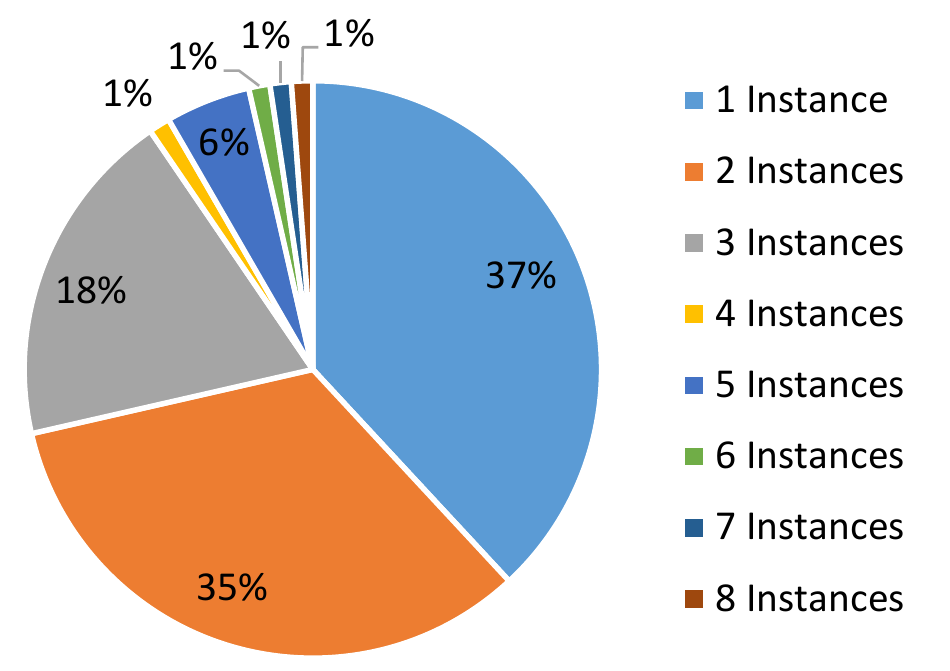} \\ {\small (a) Number of instances.} &
        \centering \includegraphics[width=1\linewidth]{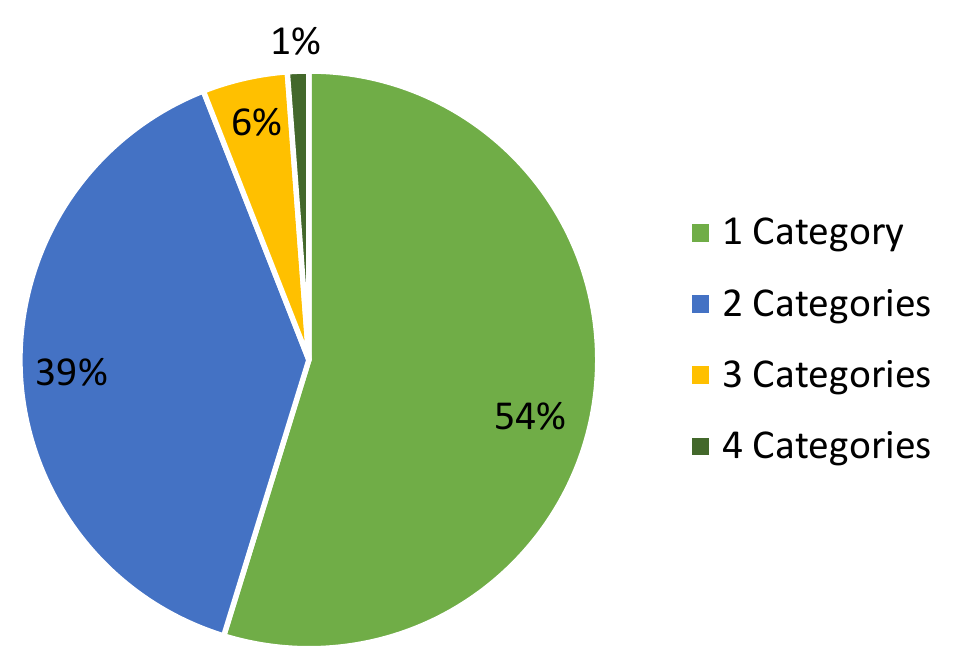} \\ {\small (b) Number of categories.} &
        \centering \includegraphics[width=1\linewidth]{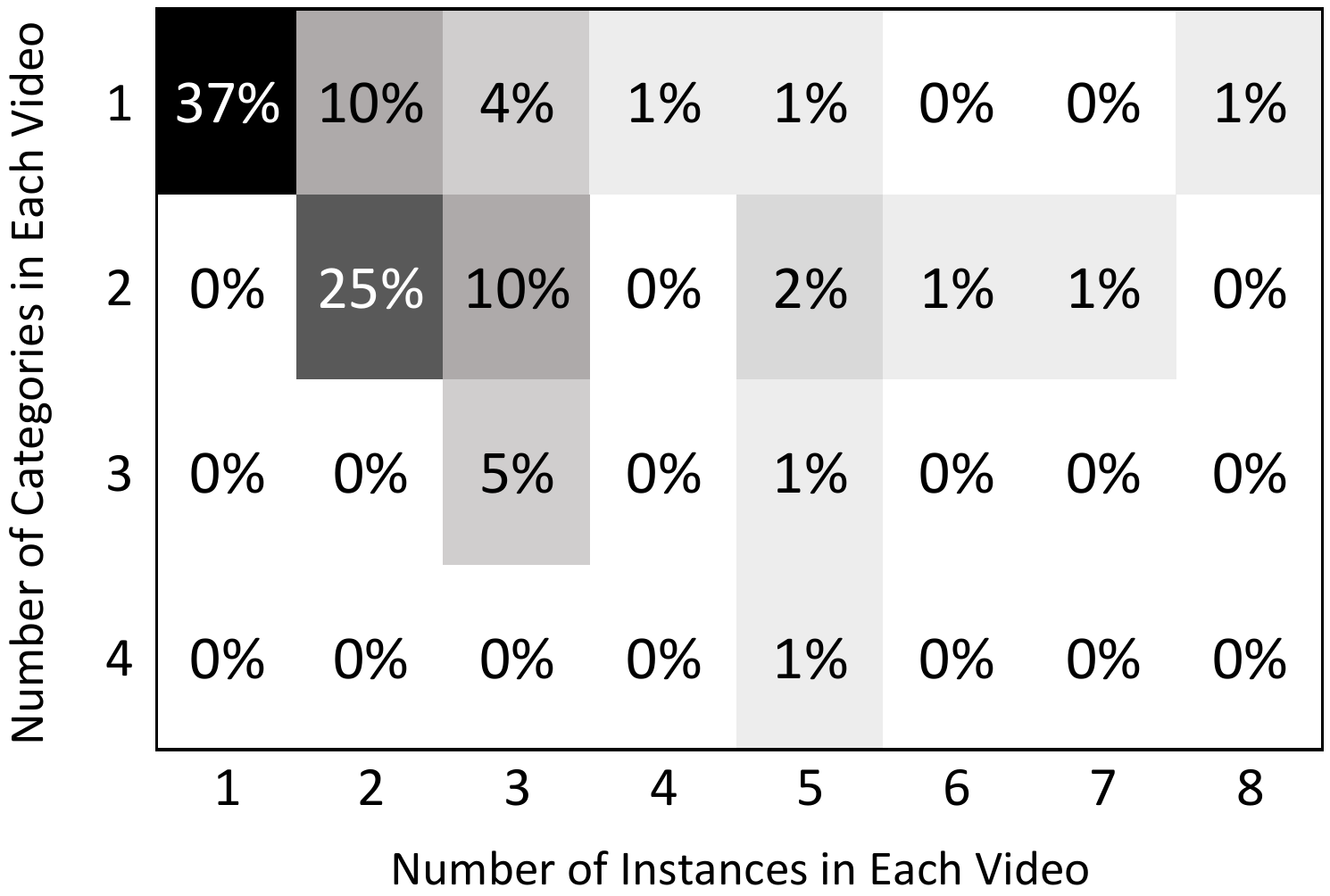} \\ {\small (c) Number of instances and categories.} 
    \end{tabularx}
    \vspace*{-.5\baselineskip}
    \caption{Statistics of videos over the SESIV dataset based on the numbers of instances/categories in each video.}
    \vspace*{-1\baselineskip}
\label{c5_fig:distribution}
\end{figure*}


To build the dataset, we used 90 videos in the DAVIS-2017 dataset~\cite{Jordi-2017}, which has pixel-wise instance-level ground-truth
. This dataset is designed for semi-supervised instance segmentation where instances are indicated in the first frame of the video regardless of whether they are salient. Therefore, instance labels in the DAVIS-2017 dataset are annotated regardless of whether they are salient or non-salient, and they do not contain any semantic labels. 

In order to construct annotations for the task of VSSIS, we need to identify salient instances and assign a semantic label to each salient instance to create the SESIV dataset. Figure~\ref{c5_fig:dataset_construction} illustrates the flowchart of constructing the SESIV dataset. We first manually eliminated non-salient instances and kept only salient instances (Fig.~\ref{c5_fig:dataset_construction} (a)). 
Then, we annotated semantic labels to the instances to have semantic salient instances using 29 among 80 categories of the MS-COCO dataset~\cite{Lin-ECCV2014}  (Fig.~\ref{c5_fig:dataset_construction} (b)).  They are \textit{person, bicycle, car, motorcycle, airplane, bus, train, truck, boat, bird, cat, dog, horse, sheep, cow, elephant, bear, backpack, snowboard, sports ball, kite, skateboard, surfboard, tennis racket, chair, tv, remote, cell phone,} and \textit{clock}. The famous MS-COCO dataset is the largest large-scale dataset for object detection/segmentation to the date, thus synchronizing our SESIV with MS-COCO has advantages for training models. After that, we merged unlabeled salient instances into their neighboring one so that the merged instance can be labeled. For example, "mask" instance is merged into "clothes" instance to obtain a new instance that is annotated with person (Fig.~\ref{c5_fig:dataset_construction} (c)). 
Finally, we discarded six videos, namely, \textit{camel, goat, gold-fish, pigs, rhino,} and \textit{varanus-cage} as in Fig.~\ref{c5_fig:dataset_construction} (d), because these videos do not have any labeled semantic salient instances.

\subsection{Dataset Description}

The SESIV dataset consists of 84 videos, and the average length of the 84 videos is 68 frames. 
We note that $28\%$ of the videos have from 71 to 80 frames.
We also note that the challenge of the SESIV dataset is enhanced due to the same properties as the DAVIS-2017 dataset~\cite{Jordi-2017}.
They are \textit{background clutter, dynamic background, deformation, appearance change, shape complexity, small instance, occlusion, out of view, motion blur,} and \textit{fast motion}.

We here analyze in-depth two other properties that are specifically designed for VSSIS:
\begin{itemize}
    \item Number of semantic salient instances.
    \item Number of categories used for semantic annotation.
\end{itemize}
We present the distribution of these two properties over the SESIV dataset in Fig.~\ref{c5_fig:distribution}. 
Each video has the maximum of 8 semantic salient instances. 
Most videos have from 1 to 3 semantic salient instances: $37\%$ of the videos have one instance, $35\%$ do two instances, and $18\%$ do three instances (Fig.~\ref{c5_fig:distribution} (a)). 
Each video has the maximum of 4 categories. 
$54\%$ of the videos have only one category while $39\%$ do two categories (Fig.~\ref{c5_fig:distribution} (b)). 
A large number of videos have a single instance (37\%) or two instances from different categories (25\%) (Fig.~\ref{c5_fig:distribution} (c)). 


It is also noteworthy that instances can disappear in several frames in a video due to, for example, \textit{full occlusion} or \textit{out of view}. 
$17.9\%$ of the videos have instances that disappear in their some frames.
They are, for example, ~\textit{bmx-bumps, color-run, dog-gooses, drone, surf}, and \textit{walking}.



\section{Proposed Baseline}
\label{c5_section:method}

\subsection{Overview}

The most straightforward approach to VSSIS is to segment individual instances frame-by-frame and then combine them to obtain final results. However, this approach does not guarantee consistency of labels over frames due to frame-by-frame processing. Furthermore, this approach faces the problem that instances overlap with each other.

To overcome such issues, we propose a \textbf{S}emantic \textbf{I}nstance - \textbf{S}alient \textbf{O}bject (SISO) baseline, consisting of two streams (one for SIS, and the other for SOS), where salient instances in the current frame are propagated to those in subsequent frames to maintain consistency of their labels, in terms of identity and semantic, even if instances disappear in some frames. Therefore, SISO is able to deal with a varying number of salient semantic instances and is scalable to the length of videos.

\begin{figure}[t]
    \centering
        \includegraphics[width=1.1\linewidth]{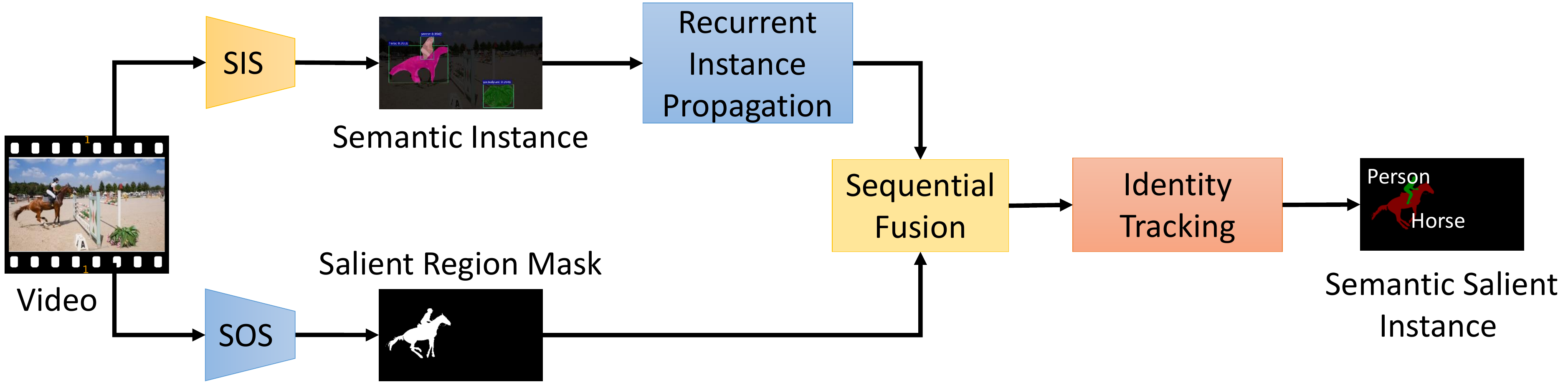}
    \caption{Pipeline of proposed SISO baseline. Yellow, orange, and blue blocks are for spatial, temporal, and spatiotemporal computation, respectively. Both two streams can work on the spatiotemporal domain.}
    \vspace*{-1\baselineskip}
    \label{c5_fig:SISO}
\end{figure}

Figure~\ref{c5_fig:SISO} illustrates pipeline of SISO. Two streams (\eg~ SIS and SOS) of SISO work on both spatial and temporal domains. Outputs of streams are fused to remove non-salient instances, producing a pixel-wisely labeled instance map. We remark that both salient region mask and semantic instances are spatially refined before the fusion, using boundary snapping method~\cite{Caelles-CVPR2017, ltnghia-DAVIS2017, ltnghia-BMVC2017}, improving accuracy of the final result. Finally, the identity tracking maintains the consistency of instance labels over the entire video.

\textbf{SIS Stream.} No existing work can deal with SIS in videos. We thus use the strategy of frame-by-frame segmentation followed by instance linkage over the entire video. Particularly, semantic instances segmented at each frame are temporally propagated over the entire video using the recurrent instance propagation to improve the accuracy of instance shapes. In section \ref{c5_section:experiment_baseline}, we evaluate the performance of various network architectures implemented in the SIS stream.

\textbf{SOS Stream.} We employ the 3D FCN model proposed by Le~\etal\cite{ltnghia-BMVC2017} as the SOS stream, thus computed saliency map implicitly contains temporal information. The saliency map is then binarized to have salient region mask using, for example, the adaptive threshold $\theta=\mu+\eta$ where $\mu$ and $\eta$ are the mean value and the standard deviation of pixel-wise saliency values over the frame.



\subsection{Sequential Fusion}

\begin{algorithm}[t]
    \caption{Sequential fusion and confidence computation.}
    \label{c5_alg:sequential_fusion}
    {\small
    \SetAlgoLined
    \SetKwInOut{Input}{input}\SetKwInOut{Output}{output}\SetKwInOut{Parameter}{parameter}
    \Input{salient region mask $M$; set of instance identities $I$, where each instance $i \in I$ has region $R_i$, category $C_i$, and classification score $S^{({\rm cls})}_i$}
    \Output{fusion map $FM$; frame-confidence $FC$, set of confident-instance identities $J$, where each instance $j \in J$ has new region $\widetilde R_j$}
    \Parameter{threshold $\theta$}
    \BlankLine
    \emph{initialize:} $FM \leftarrow [0]^{h \times w}$; 
    $S^{({\rm conf})} \leftarrow 0$; 
    $J \leftarrow \emptyset$\;
    \Repeat{$\max S^{({\rm seg})} \leq \theta$ {\bf or} $S^{({\rm seg})} == \emptyset$}{
        $S^{({\rm seg})} \leftarrow [0]^{|I|}$;   \tcp{\footnotesize {set of segmentation scores}}
        \For{$i \in I$}{
            $S^{({\rm seg})}_i \leftarrow$ {\bf IOU} $(R_i, M)$\;
            \If{$S^{({\rm seg})}_i == 0$}{
                $S^{({\rm cls})}_i \leftarrow \emptyset$; 
                $S^{({\rm seg})}_i \leftarrow \emptyset$; 
                $R_i \leftarrow \emptyset$; 
                $C_i \leftarrow \emptyset$\;
            }
        }
        $j \leftarrow \arg \mathop {\max }\limits_i S_i^{({\rm seg})}$\;
        $\widetilde R_j \leftarrow R_j \cap M$\;
        Pixels in $FM$ corresponding to $\widetilde R_j \leftarrow C_j$\;
        $M \leftarrow M \backslash \widetilde R_j$; $J \leftarrow J \cup \{j\}$\;
        $S^{({\rm conf})} \leftarrow S^{({\rm conf})} +$ {\bf CS}$(S^{({\rm seg})}_j, S^{({\rm cls})}_j)$\;
        $S^{({\rm cls})}_j \leftarrow \emptyset$; 
        $S^{({\rm seg})}_j \leftarrow \emptyset$; 
        $R_j \leftarrow \emptyset$; 
        $C_j \leftarrow \emptyset$\;
    }
    $FC \leftarrow \frac{S^{({\rm conf})}}{|J|}$
    }
\end{algorithm}

When fusing semantic instances with the salient region mask for the semantic salient instance map, dealing with the areas where different instances overlap with each other becomes a crucial issue. 

Almost all multi-instance segmentation methods ignore such areas and randomly merge instances~\cite{Cheng-DAVIS2017, Newswanger-DAVIS2017, Shaban-DAVIS2017, Sharir-DAVIS2017, Voigtlaender-DAVIS2017}. Though Le \etal~\cite{ltnghia-DAVIS2017} proposed to merge instances depending on the order based on their topological relationships, 
their method requires the ground-truth label of the first video frame to learn the order. 
We here propose a novel sequential fusion that does not require any ground-truth label. 
We compute the merging order in each frame using the salient region mask from the SOS stream. 

Algorithm~\ref{c5_alg:sequential_fusion} describes our proposed sequential fusion to select a set of instances (hereafter, referred to confident-instances) to compute a fusion map. 
We select the instance that overlaps the salient region mask best 
where we use IOU~\cite{Hariharan-ECCV2014} to compute the overlapping area between the instance and the mask. We set semantic label of the selected instance to each pixel in its corresponding region of the fusion map. 
We then remove the overlapping area from the salient region mask. 
Next, we select the instance from other remaining instances that overlaps the remaining salient region mask best and then remove the overlapping area.
We iterate this procedure until no instance exists inside the remaining salient region mask.  
In our experiments, when the IOU score for an instance is less than $\theta=0.1$, we regarded the instance is not present in the salient region mask. 


We also compute the frame-confidence for each frame by averaging the confidence scores of all the semantic salient instances in the frame. 
The confident score of a semantic salient instance, denoted by ${\bf CS}(\cdot)$, is computed 
as a trade-off between the IOU score and the classification accuracy: 
${\bf CS}=\frac{(1+\beta^2)S^{({\rm seg})}S^{({\rm cls})}}{\beta^2S^{({\rm seg})}+S^{({\rm cls})}}$, where 
$S^{({\rm seg})}$ is the segmentation IOU score of the instance and $S^{({\rm cls})}$ is the classification accuracy score of the instance. 
We remark that in our experiments we set $\beta^2=0.3$ so that the segmentation score $S^{({\rm seg})}$ is more weighted. 

\subsection{Recurrent Instance Propagation}

 \begin{figure}[t]
     \centering
         \includegraphics[width=1\linewidth]{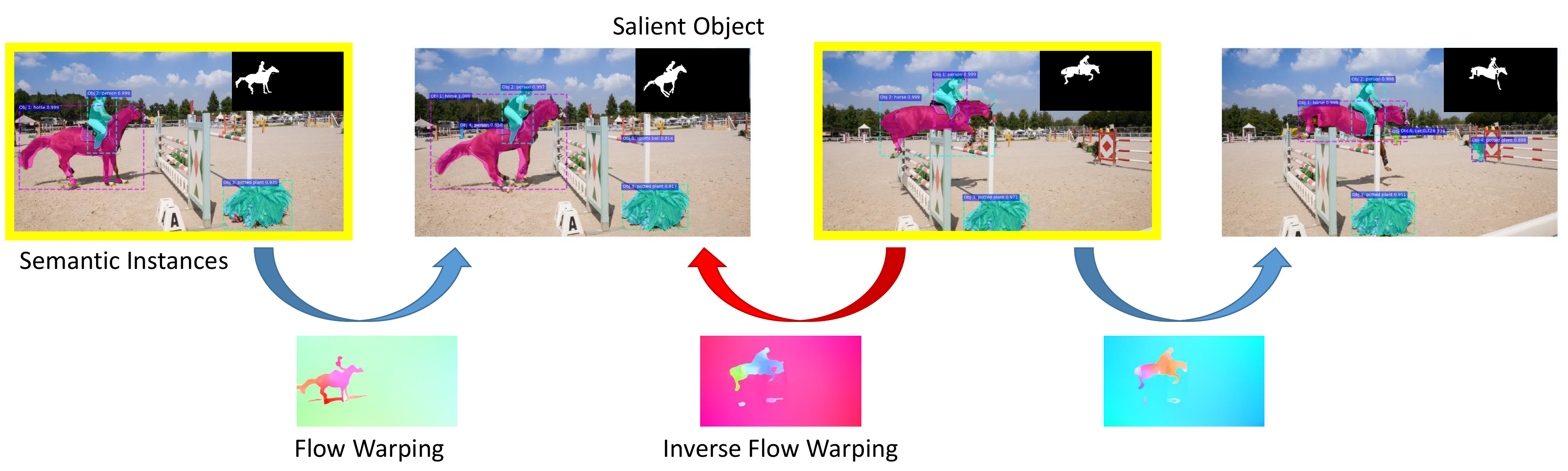}
     \vspace*{-1.5\baselineskip}
     \caption{Flowchart at one iteration of the recurrent instance propagation. Semantic instances are propagated from video frames with high frame-confidences to ones with low frame-confidences. Video frames with a yellow bounding box have higher frame-confidences than their adjacent frames.}
     \label{c5_fig:mask_propagation}
 \end{figure}

Some semantic instances may not be segmented due to severely deformed appearances caused by object motion and/or camera motion. 
To recover such missing semantic instances, we introduce the recurrent instance propagation where instances are recurrently propagated to neighboring frames. 
More specifically, We propagate instances in frames with high frame-confidences to those in frames with low frame-confidences. 

Figure \ref{c5_fig:mask_propagation} depicts the flowchart of the propagation at one iteration. 
Video frames are first sorted in the descending order based on their frame-confidences computed by Algorithm~\ref{c5_alg:sequential_fusion}. 
We sequentially update confident-instances and frame-confidences of all video frames in this order. 
If a video frame has larger frame-confidence than its adjacent frames, instances of the frame are propagated to the next frame and the previous frame using flow wrapping/inverse flow wrapping where the flow is computed using FlowNet2~\cite{Ilg-CVPR2017}. 
The propagated instances are then integrated to instances already segmented in the target frame. 
After that, we re-compute frame-confidence and confident-instances of the target frame. 
If the frame-confidence increases, we update the frame-confidence and confident-instances of the target frame. 
After updating frame-confidences of all the video frames, the average confidence of the video is computed by averaging all frame-confidences. 
This propagation is recurrently executed until the average confidence of the video converges. 
We remark that we empirically observe that semantic salient instances are effectively propagated after around five iterations. 


\subsection{Identity Tracking}

\begin{figure}[t]
    \centering
        \includegraphics[width=1\linewidth]{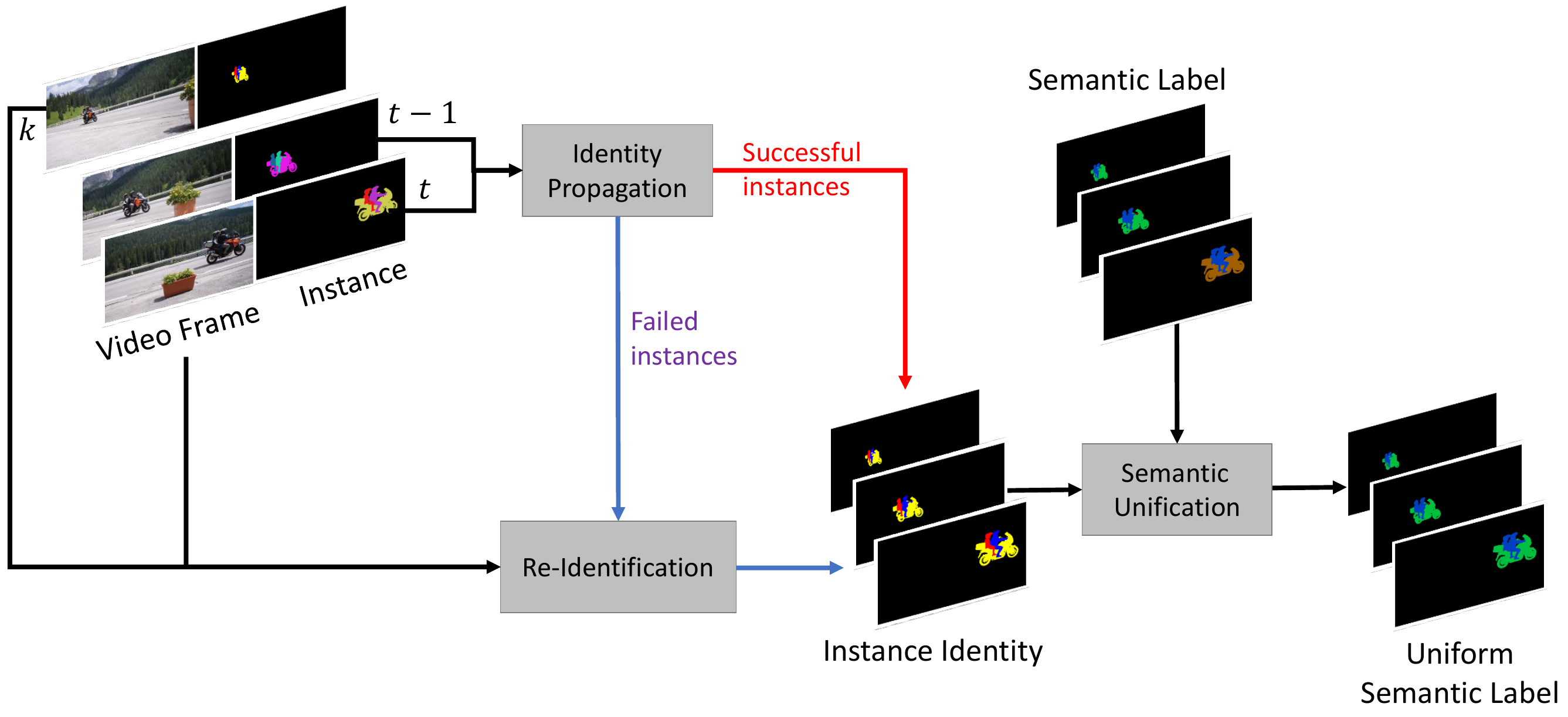}
    \vspace*{-1.5\baselineskip}
    \caption{Flowchart of identity tracking module. Instance identities from a video frame are propagated to its new frames by flow warping. When an instance is occluded or out of frame, it is re-identified in next frames by using the feature extracted at its key-frame $k$. The consistency of identities and semantic labels of instances is maintained over the entire video.}
    \vspace*{-1\baselineskip}
    \label{c5_fig:instance_propagation}
\end{figure}

Since the semantic label of an instance is attached frame-by-frame, how to maintain the consistency of the label over the entire video is critical. 
To enable SISO to maintain this consistency, we introduce the identity tracking
where the identities of instances are propagated over frames to maintain short-term consistency and
they are re-identified and unified for long-term consistency.
With this identity tracking, the identities of instances are consistently tracked over the entire video even if the instances disappear (or are occluded) and re-appear in some frames in the video.
Fig.~\ref{c5_fig:instance_propagation} depicts the flowchart of our proposed identity tracking. 
 


\subsubsection{Identity Propagation}

We initialize identities of instances in the first frame. 
The identity propagation propagates the identifies of instances in a given frame to its next frame using flow warping.
We then check how each propagated instance overlaps with instance already segmented in the target frame.
Namely, for a propagated instance, we compute IOU~\cite{Hariharan-ECCV2014} scores between the instance and each of the instances already segmented in the target frame.  
We then update the identity of the instance having the largest IOU score so that it is the same with the identity of the propagated instance.
If none of the instances in the target frame achieves $\theta=0.7$ of the IOU score, we regard that the propagated instance is out of frame or occluded in the target frame.
Re-identification is required for such an instance.

We note that any instance at the target frame that is not propagated from the previous frame is regarded as a new instance and annotated with a new identity.


\subsubsection{Re-identification}

We employ instance search~\cite{Salvador-CVPRW2016} for re-identifying instance identity, where we use feature of an instance of interest in a previous frame to detect the instance in future frames.

Given an instance of interest to be re-identified in a target frame, we first select its key-frame from previous frames and then extract a query feature from the bounding-box around the region of the instance in the key-frame. 
After that, we apply Faster R-CNN~\cite{Ren-NIPS2015} to the target frame to generate region proposals and extract features from each of proposed regions. 
We then select the proposed region that is most similar to the instance based on cosine similarity between the query feature and the feature extracted from each region. 
Next, we compute IOU~\cite{Hariharan-ECCV2014} between the selected proposed region and each region of all instances already segmented in the target frame. 
If the largest IOU score is larger than the threshold $\theta=0.7$, the corresponding instance is updated with identity of the instance of interest. 

For instance $i$, the key-frame is selected as follows.
The instance may have multiple separated regions in a frame.
We thus compute the average area of connected regions of the instance $i$ in a frame $t$:
$S^{\rm (area)}_{i,t}=\frac{area_{i,t}}{n_{i,t}}$,  where 
$area_{i,t}$ is the area where instance $i$ exists at frame $t$, and
$n_{i,t}$ denotes the number of separated regions of instance $i$ at frame $t$. 
The key-frame of the instance $i$ is given by $\arg \mathop {\max }\limits_t S^{\rm (area)}_{i,t}$.

\subsubsection{Semantic Unification}


For a semantic salient instance and a category, we first compute the summation over the entire video of the classification scores that
the instance belongs to the category. We then choose for the instance the semantic label of the category that achieves the maximum value among all the categories. In this way, the semantic labels attached to salient instances are unified over the entire video. 




\begin{table}[t]
\centering
\caption{Results on SESIV. The best results are shown in \textcolor[rgb]{0,0,1}{blue}.}
\vspace*{-.5\baselineskip}
\label{c5_tab:result}
\small
\begin{tabular}{l|rr}
\toprule
\centering \textbf{Method} & $\mathcal{JS}$ & $\mathcal{FS}$ \\
\midrule
{Mask R-CNN$_{\rm org}$}~\cite{Kaiming-ICCV2017} & 0.41 & 0.43 \\
{Mask R-CNN$_{\rm prop}$} & 15.65 & 16.70 \\
{Mask R-CNN$_{\rm SISO}$} & \textcolor[rgb]{0,0,1}{41.71} & \textcolor[rgb]{0,0,1}{42.59} \\
\midrule
{MNC$_{\rm org}$}~\cite{Dai-CVPR2016} & 0.72 & 0.72 \\
{MNC$_{\rm prop}$} & 20.36 & 18.97 \\
{MNC$_{\rm SISO}$} & \textcolor[rgb]{0,0,1}{31.07} & \textcolor[rgb]{0,0,1}{31.08} \\
\bottomrule
\end{tabular}
\end{table}


\section{Experiments} 
\label{c5_section:setting}


\subsection{Implementation Details}
\label{c5_section:implementation}


For the SOS stream, we employed DSRFCN3D~\cite{ltnghia-BMVC2017}, using the public pre-trained model on video saliency datasets~\cite{ltnghia-BMVC2017} (without any fine-tuning).

For the SIS stream, we employed Mask R-CNN~\cite{Kaiming-ICCV2017} and MNC~\cite{Dai-CVPR2016} to evaluate the performance of the proposed SISO on various network architectures. We used public pre-trained models without any fine-tuning (Mask R-CNN is pre-trained on the MS-COCO dataset~\cite{Lin-ECCV2014}, and MNC is pre-trained on the VOC Pascal dataset~\cite{Everingham-ICCV2010}). 
We remark that we used only semantic instances whose classification scores are larger than 0.7; we eliminated the other instances. We also remark that to evaluate MNC, we converted semantic ground-truth labels of the MS-COCO to their corresponding categories of the VOC Pascal and used only convertible semantic salient instances. 

We implemented optical flow~\cite{Ilg-CVPR2017}, instance search~\cite{Salvador-CVPRW2016}, and SIS models~\cite{Dai-CVPR2016, Kaiming-ICCV2017} with python, VSOS model~\cite{ltnghia-BMVC2017} and other modules with Matlab. All experiments were conducted on a computer with a Core i7 3.6GHz processor, 32GB of RAM, and GTX1080Ti GPU.

\subsection{Evaluation Criteria}
\label{c5_section:metrics}



To evaluate performances, we introduce semantic region similarity and semantic contour accuracy defined as follows. Let $m$ and $g$ be binary masks of the predicted instance and the ground-truth instance. 
The semantic region similarity $\mathcal{JS}$ and the semantic contour accuracy $\mathcal{FS}$ are
\begin{eqnarray}
\mathcal{JS}(m,g)=\delta_{id(m), id(g)}\delta_{sl(m), sl(g)}\mathcal{J}(m,g), \\
\mathcal{FS}(m,g)=\delta_{id(m), id(g)}\delta_{sl(m), sl(g)}\mathcal{F}(m,g),
\end{eqnarray}
where $\mathcal{J}(\cdot)$ and $\mathcal{F}(\cdot)$ are region similarity~\cite{Everingham-ICCV2010} and contour accuracy~\cite{Perazzi-CVPR2016}. 
$\delta$ denotes the Kronecker delta, and $id(m)$ and $sl(m)$ are the identity and the semantic label of instance $m$, respectively. 
Remark that we compare the similarity of two instances only if they have the same identity and the same semantic label. 
We note that region similarity is the intersection over the union of the estimated segmentation and the ground-truth mask while contour accuracy is a trade-off between the contour-based precision and recall. 

\begin{table}[t]
\centering
\caption{Effectiveness of confident instance utilization. The best results are shown in \textcolor[rgb]{0,0,1}{blue}.}
\vspace*{-.5\baselineskip}
\label{c5_tab:instance_fusion_effectiveness}
\resizebox{\linewidth}{!}{%
\small
\begin{tabular}{l|c|c|c}
\toprule
\centering \textbf{Method or Metric} & \textbf{\textit{$SISO_a$}} & \textbf{\textit{$SISO_b$}} & \textbf{\textit{$SISO_c$}} \\
\midrule
Sequential Fusion &  & \checkmark & \checkmark \\ 
Recurrent Instance Propagation &  &  & \checkmark  \\
\midrule
$\mathcal{JS}$ & 36.57 & 37.43 & \textcolor[rgb]{0,0,1}{41.71} \\
$\mathcal{FS}$ & 39.59 & 40.32 & \textcolor[rgb]{0,0,1}{42.59} \\
\bottomrule
\end{tabular}
}
\end{table}

\begin{table}[t]
\centering
\caption{Effectiveness of identity tracking. The best results are shown in \textcolor[rgb]{0,0,1}{blue}.}
\vspace*{-.5\baselineskip}
\label{c5_tab:instance_propagation_effectiveness}
\small
\begin{tabular}{l|c|c|c}
\toprule
\centering \textbf{Method or Metric} & \textbf{\textit{$SISO_{\alpha}$}} & \textbf{\textit{$SISO_{\beta}$}} & \textbf{\textit{$SISO_{\gamma}$}} \\
\midrule
Identity Propagation &  & \checkmark & \checkmark \\ 
Re-Identification &  &  & \checkmark  \\
\midrule
$\mathcal{JS}$ & 0.95 & 33.74 & \textcolor[rgb]{0,0,1}{41.71} \\
$\mathcal{FS}$ & 1.02 & 34.56 & \textcolor[rgb]{0,0,1}{42.59} \\
\bottomrule
\end{tabular}
\end{table}

Similar to \cite{Jordi-2017}, we first evaluate each instance and then take the average over the dataset. 
More precisely, letting $V$ be a set of videos in the dataset, and $\mathcal{M} \in \{\mathcal{JS}, \mathcal{FS}\}$ be a given metric,
the performance $\mathcal{M}(V)$ over $V$ is defined by
\begin{eqnarray}
\mathcal{M}(V) = \frac{1}{|I_V|}\sum\limits_{i \in I_V}{\frac{1}{|F_{v(i)}|}\sum\limits_{f \in F_{v(i)}}{\mathcal{M}(m^f_i, g^f_i)}},
\end{eqnarray}
where $I_V$ is the set of annotated instances in $V$, $v(i) \in V$ is the sequence in which the instance $i \in I_V$ appears, and $F_v$ is the set of frames in sequence $v$.
$m^f_i$ and $g^f_i$ are respectively the predicted region and the ground-truth of instance $i$ in frame $f$.


We remark that we matched identities of predicted instances at the first frame with those of the ground-truth by maximizing IOU scores between the predicted instances and the ground-truth. This avoids the identity permutation problem in the evaluation.



\subsection{Results of SISO Instances}
\label{c5_section:experiment_baseline}



We emphasize that SISO is the first work for VSSIS, meaning that no state-of-the-art method is available for comparison. We thus evaluated the performance of various network architectures implemented in the SIS stream. Each method $M$, where $M=\{$Mask R-CNN, MNC$\}$, is employed with three different settings: $M_{\rm org}$ is the original model (we applied this frame-by-frame for videos), $M_{\rm prop}$ is the model incorporating our identity propagation module (this is just to simply exploit temporal information), and $M_{\rm SISO}$ is the model incorporated in our proposed SISO.

\begin{figure*}[t]
    \centering
        \includegraphics[width=1\linewidth]{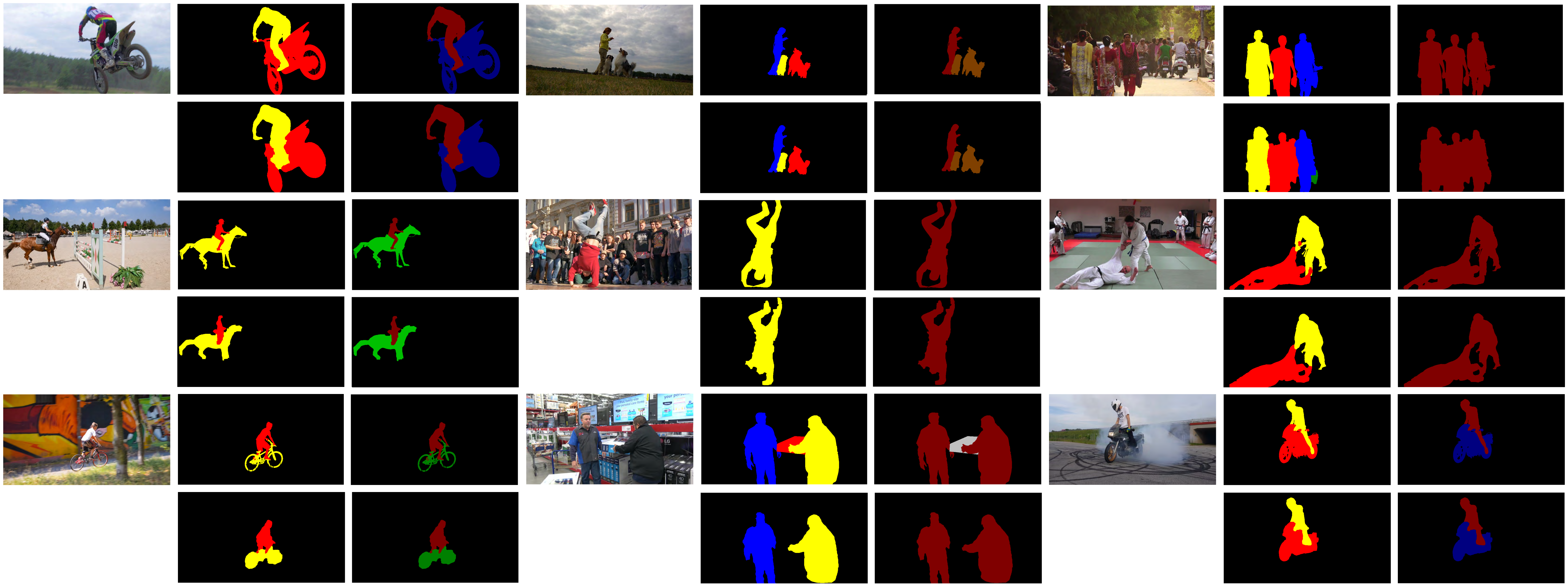}
    \caption{Visualization of some results by our method on the SESIV dataset. From left to right, original video frame is followed by instance label and semantic label, respectively. The top row indicates ground-truth labels and the bottom row shows results by our method.}
    \label{c5_img:visual_comparison_sesiv}
\end{figure*}

The quantitative results are shown in Table~\ref{c5_tab:result}, indicating that SISO significantly outperforms the other settings for any SIS method on all metrics. This suggests that SISO is capable of eliminating non-salient instances and maintaining consistent identities of instances over the entire video. We also note that the setting $M_{\rm org}$ achieves the worst performances. This is because it is a frame-by-frame method and does not take into account temporal information. Figure \ref{c5_img:visual_comparison_sesiv} is the visualization of a few examples obtained by Mask R-CNN$_{\rm SISO}$. We see that our method handles complex instances with background clutter, giving accurate and consistent segmentation. 


\subsection{Ablation Studies}


To demonstrate the effectiveness of components in SISO, i.e., sequential fusion, recurrent instance propagation, and identity tracking, we performed experiments under controlled settings and compared results. We note that we used Mask R-CNN$_{\rm SISO}$ for these experiments because we see that Mask R-CNN$_{\rm SISO}$ performed better than MNC$_{\rm SISO}$.


\subsubsection{Effectiveness of Confident-Instance Utilization}


As shown in Section~\ref{c5_section:method}, confident-instances are utilized in the sequential fusion and the recurrent instance propagation modules. To evaluate the effectiveness of confident-instances, we performed experiments under three different controlled settings: merging instances in the random order without using any confident-instance (denoted by $SISO_{a}$), using the sequential fusion only (denoted by $SISO_{b}$), and using both the sequential fusion and the recurrent instance propagation (denoted by $SISO_{c}$). Table \ref{c5_tab:instance_fusion_effectiveness} shows their results, indicating that (1) merging instances based on introduced confident-instances ($SISO_{b}$) achieves better performance than in the random order ($SISO_{a}$), and that (2) utilizing confident instances ($SISO_{b}$ and  $SISO_{c}$) performs better than not using confident-instances ($SISO_{a}$). In particular, our complete method $SISO_{c}$ performs best. 



\subsubsection{Effectiveness of Identity Tracking}

\begin{figure*}[t]
     \centering
     \includegraphics[width=1\textwidth]{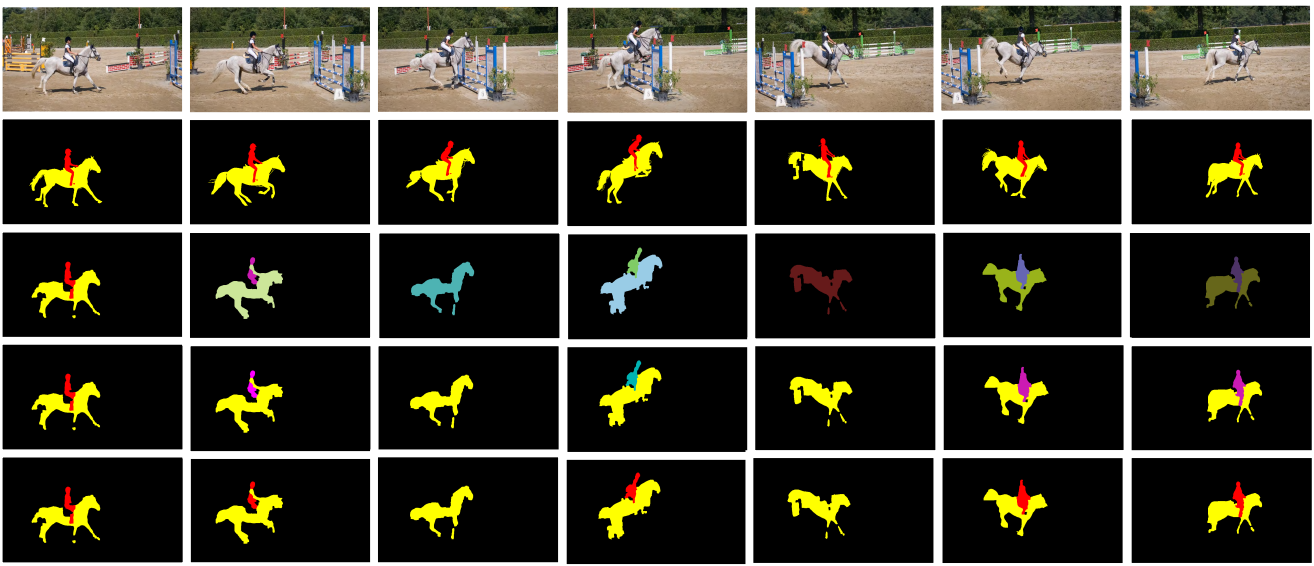}
     \caption{Effectiveness of identity tracking in the temporal domain. From top to bottom, original video frame and ground-truth are followed by outputs obtained by different settings: no tracking instances, performing only identity propagation, and performing identity tracking (including both identity propagation and re-identification).}
     \label{c5_fig:instance_propagation_effectiveness}
 \end{figure*}

To evaluate the effectiveness of the identity tracking module, we performed experiments under three different controlled settings: not tracking any instances (denoted by $SISO_{\alpha}$), using the identity propagation only (denoted by $SISO_{\beta}$), and using both the identity propagation and the re-identification (denoted by $SISO_{\gamma}$). Evaluation results are illustrated in Fig.\ref{c5_fig:instance_propagation_effectiveness}.
Table \ref{c5_tab:instance_propagation_effectiveness} shows their results and indicates that our complete method $SISO_{\gamma}$ exhibits outperformance against the other settings on all the metrics. In particular, the outperformance over $SISO_{\alpha}$ is significant. We also observe that using both the identity propagation and the re-identification ($SISO_{\gamma}$) brings more gains than using the identity propagation only ($SISO_{\beta}$). This suggests that the identity tracking contributes to maintain consistent identities of instances over the entire video.


\section{Conclusion}
\label{c5_section:conclusion}


We addressed a new task of video semantic salient instance segmentation (VSSIS), and proposed the first baseline for for VSSIS, called Semantic Instance - Salient Object (SISO). SISO is a simple yet efficient framework that jointly performs semantic instance segmentation and salient object segmentation in videos. Furthermore, SISO is capable of eliminating non-salient instances and maintaining consistency of both identities and semantic labels for salient instance over the entire video thanks to our introduced sequential fusion, recurrent instance propagation, and identity tracking. 
To address the task of VSSIS, we provided a new dataset SESIV consisting of 84 video sequences with pixel-wisely annotated per-frame ground-truth labels. 

Besides extending the quantity of the dataset, developing a way to directly segment salient instances from videos is 
left for future work.

{\small
\bibliographystyle{ieee}
\bibliography{shortbib}
}

\end{document}